\pdfoutput=1
\relax
\documentclass[letterpaper]{article}
\usepackage{times}
\usepackage{helvet}
\usepackage{courier}
\usepackage{url}
\usepackage{graphicx}
\title{$FairMod$ - Making Predictive Models Discrimination Aware}
\author{Jixue Liu$^{a,1}$, Jiuyong Li$^{a}$, Lin Liu$^{a}$, Thuc Duy Le$^{a}$, Feiyue Ye$^{b,1}$, Gefei Li$^{a}$\\
$^a$ University of South Australia  \hspace{3em}  $^b$ Jiangsu University of Technology, China\\
\{jixue.liu,jiuyong.li,lin.liu,thuc.le,gefei.li\}@unisa.edu.au; yfy@jsut.edu.cn
}
\usepackage{booktabs} % For formal tables
\usepackage{enumerate}
\usepackage{verbatim}

\usepackage{algorithm,algorithmic,  xcolor}
\usepackage[fleqn]{amsmath}
\usepackage{times, epsfig, theorem, amssymb,graphicx, wrapfig}

\usepackage[square, numbers, sort]{natbib}

\newtheorem{definition}{Definition}

\newtheorem{lemma}{Lemma}

\renewcommand{\emph}{\textit}

\newcommand{\ttop}[1]{\ensuremath{\hspace{-.2ex}{{\small #1}}\hspace{-.2ex}}}
\newcommand{\vv}[1]{\avlong#1!}
\def\avlong#1,#2!{{#1=#2}}
\newcommand{\mb}[1]{\ensuremath{\mathbf{#1}}}
\newcommand{\mbE}{\ensuremath{\mathbf{E}}}

\newcommand{\mbP}{\ensuremath{\mathbf{P}}}

\newcommand{\mbO}{\ensuremath{\mathbf{O}}}

\newcommand{\calM}{\ensuremath{{\cal M}}}

\newcommand{\bfit}[1]{\textbf{\textit{#1}}}

\newcommand{\bref}[1]{(\ref{#1})}

\newcommand{\smallpara}[1]{\vspace{.5ex} \noindent {\bf #1}}
\newcommand{\smallsubsec}[1]{\vspace{1ex}\noindent {\bf\large #1}\\\vspace{0.5ex}}

\renewenvironment{thebibliography}[1]{%
    \begin{oldthebibliography}{#1}%
      \setlength{\parskip}{0ex}%
      \setlength{\itemsep}{0ex}%
  }%
  {%
    \end{oldthebibliography}%
  }

\begin{document}
\maketitle
\footnotetext[1]{supported by Grant 61472166 of National Natural Science Foundation China}
\begin{abstract}
Predictive models such as decision trees and neural networks may produce discrimination in their predictions. This paper proposes a method to post-process the predictions of a predictive model to make  the processed predictions non-discriminatory. The method considers multiple protected variables together. Multiple protected variables make the problem more challenging than a simple protected variable. The method uses a well-cited discrimination metric and adapts it to allow the specification of explanatory variables,  such as position, profession, education, that describe the contexts of the applications. It models the post-processing of predictions problem as a nonlinear optimization problem to find best adjustments to the predictions so that the discrimination constraints of all protected variables are all met at the same time. The proposed method is independent of classification methods. It can handle the cases that existing methods cannot handle: satisfying multiple protected attributes at the same time, allowing multiple explanatory attributes, and  being independent of classification model types. An evaluation using four real world data sets shows that the proposed method is as effectively as existing methods, in addition to its extra power.  
\end{abstract}

%\begin{IEEEkeywords}
%fairness computing, discrimination aware; prediction adjustment model
%\end{IEEEkeywords}

\section{Introduction}

Discrimination means ``treating a person or particular group of people differently, especially in a worse way from the way in which you treat other people, because of their skin colour, sex, sexuality, etc'' (dictionary.cambridge.org). It can happen in law enforcement applications where  people may be unfairly treated and sentenced because of their races and religions, in bank loan applications where people may not get a loan because they live in a suburb with a lower economic status. As the definition indicates, discrimination can happen to an individual or a group. We are interested in group discrimination in this paper. 

The law does not allow discrimination to happen. That is, decisions should not be made based on people's sex, skin colour, religion etc, called protected groups. However discrimination is still a concern in the real world. In the car insurance industry, insurance companies  often require  people in a specific suburb to pay higher premium than those in other suburbs with the reason that the suburb has a higher claim rate.  If most dwellers of the suburb are of a certain race, the higher premium forms actually discrimination to the people of this race. More examples can be found in \cite{threelendingDsc}. 

Discrimination also becomes a concern in automated decision making systems as machine learning and data analytics are used more and more in real life applications. This type of discrimination is called algorithmic discrimination or fair computing \cite{Hajian2016}. In loan approval applications, it is a normal practice that an automated system is used to score customer's applications. The system would have classification rules learnt from training data. Whether the rules are fair and whether the outcomes (predictions) from the classification rules are fair determine the fairness of the system. As the investigation of fairness of the predictions of such a system is difficult because individuals see only their own outcome but not others, and comparisons are not possible. Consequently it is important to ensure the fairness of the classification rules and predictions.  

The contexts of applications play an important role in algorithmic discrimination investigation and assessment. If an investigation is on whether female employees are less paid, the professions and positions held by the employees must be considered. A fair comparison between the payments of female employees and male employees must assume that the employees hold same professions and same positions. Without a context, the comparison may be between the income of a group of female CEOs with the income of a group of male kitchen hands. Using different contexts, we get different discrimination scores. We use an example to show this point.   

\noindent {\it Example 1}
Table \ref{ex-1} shows the tuple frequencies of a data set. The outcome attribute $D$ is income with domain values 1=high and 0=low, and the protected attribute is $Sex$. Without using a context attribute in Part (a), the discrimination score for females is 0 where the score is defined as the percentage of high income females taking away the percentage of high income males $ds(r)=P(D=1| Sex=F)-P(D=1| Sex=M)$ \cite{Calders2010}.  When a context is specified, the discrimination score should be calculated in each case of the context. Let $r_1$ and $r_2$ in Parts (b) and (c) be two subsets of $r$ when the context variable is the employment sector (Sec) with domain values of 1=public and 0=private. The discrimination scores for both sectors are 0.22 and -0.24 respectively.  

\begin{table}[h]  \begin{center} \small
\caption{Effect of a context variable \label{ex-1}}
\begin{tabular}{|l|l|l|}
  \multicolumn{3}{l}{ (a) $r$.\; $ds(r) \ttop{=} 0 $} \\
\hline
	 D & Female  & Male   \\ \hline
	1(H) &   10       &   15  \\  \hline  
	0(L) &   40      &  60  \\  \hline
\end{tabular} \\
\vspace{.5ex}
\begin{tabular}{|l|l|l|}
\multicolumn{3}{l}{ (b) $r_1$. Sec=1\; $ds(r_1) \ttop{=} 0.22 $}\\
\hline
	D  & Female & Male  \\ \hline
	1  &   9     &  3  \\  \hline  
	0  &   20  &  30   \\  \hline
\end{tabular}  \quad
\begin{tabular}{|l|l|l|}
\multicolumn{3}{l}{ (c) $r_2$. Sec=0\; $ds(r_2) \ttop{=} -0.24$}\\
\hline
	D  & Female & Male  \\ \hline
	1  &   1      & 12  \\  \hline  
	0  &   20      &  30   \\  \hline
\end{tabular}
\end{center}
\end{table}

Research work on algorithmic discrimination focused on two areas. One is discrimination detection. The other is discrimination removal. Detection  aims to assess whether discrimination exists in data. An important problem here is what metrics should be used to measure the level of discrimination. The works in this area are described in the section of Related Work. Discrimination removal is so that the final predictions are non-discriminatory. Some work has been done in this direction \cite{Calders2010,Kamishima2012,Zemel2013,Zafar2015,Woodworth2017,edward18-continuous-T,Kearns2018-subgroupfairness}.

This paper proposes a method, called FairMod, to remove discrimination from the predictions of classification models (classifiers), given multiple protected attributes and explanatory variables. The multiple protected variables are considered together and this makes the problem challenging because reducing discrimination of one protected variable like gender may raise the  discrimination of another protected variable like race. The multiple protected variables make the problem much more challenging. Our method is general solution to discrimination in predictive models as it is independent of classification algorithms. 

The paper makes the following contributions.

\begin{itemize}  \topsep=0mm  \parsep=0mm \itemsep=0mm \parskip=0cm \itemindent=0em
\item The proposed method models the discrimination removal problem with multiple explanatory variables and multiple protected variables as a constrained non-linear optimization problem. The variables are adjustments to the predictions of protected subgroups so that the adjustments satisfy the discrimination constraints of all the protected attributes at the same time. The objective function is to minimize the number of adjustments and minimize the final prediction errors. The method is not classifier dependent. It can be attached to any classification model for algorithmic discrimination correction.   

\item An evaluation of our proposed method using four real world data sets from different applications indicates that the proposed method is effective in removing discrimination from the output of a classifier. It also shows in the cases (single protected attribute, no context groups) where existing methods apply, the proposed method performs as effective as existing methods. 
\end{itemize}

%The work of this paper proposes a method for post-prediction processing, but at the same time, it is extension of the classifier. The work is innovative in a few areas. Firstly this method does not limit itself with any specific classifier. That is, it can be combined with classifier to remove discrimination in the prediction. Secondly, it allows explanatory attributes which enables discrimination to be discussed in contexts. Thirdly it adjusts prediction by systematically considering all protected attributes at the same time. 

\section{Definitions and Problem}
In this section, we define the basic notation, a discrimination metric, and the problem of the paper. 

Let $r$ be a data set on a schema $\mb{R}$ of binary attributes (variables). The attributes in $\mb{R}$ are of four types: an outcome/target attribute $D$, some protected attributes $\mbP$, some explanatory attributes $\mbE$, and the other attributes $\mb{O}$: $\mb{R}=\{D\}\cup \mbP \cup \mbE \cup \mb{O}$.  With the outcome attribute, $D=1$ means a favorite outcome like $Income=High$ or $Application=Successful$ that an individual prefers to receive. With a protected attribute $P\in \mbP$, $P=1$ (e.g. Sex=Female or Race=Black) means a group of individuals who are protected by the law not to to be discriminated. The explanatory attributes  $\mb{E}$ explain why some people receive favorite outcomes more or less frequently than others or identify such people. For example, profession is an explanatory attribute. Surgeons as a profession are high income earners, while kitchen hands are low income earners.  Examples of the {\it Other} attributes can be  living suburbs and owning a house.

An  {\it E-group} (or a stratum of $\mb{E}$) is a subset $e$ of all the tuples satisfying  $\mb{E}=\mb{e}$ on all explanatory attributes $\mb{E}$ in the data set $r$. $\mb{E}=\mb{e}$, or equivalently  $(E_1=e_1, ..., E_k=e_k)$, is called the signature of the group and is denoted by $e.sig$. The concept of an E-group is fundamental in our discrimination definition. All E-groups  in $r$ are denoted by $stra(\mb{E},r)$. Corresponding to E-groups, we also define a P-group to be all the tuples having the same $\mb{P}=\mb{p}$ value. 

We note that in \cite{Kearns2018-subgroupfairness}, the subgroup are our P-groups and their work do not consider E-groups. 

\smallpara{Discrimination score} \ \\
We employ the well cited discrimination score defined in  \cite{Calders2010} and score is $Pr(D=1|\vv{P,1})   -Pr(D=1|P=0)$. In the case where $D$ is income and $P$ is gender, the score reflects the probability difference of high income earners caused by gender difference. Considering E-groups, the score for each E-group is: 
\begin{align}
&&\delta(P,\mb{e}) & =  Pr(D=1|\vv{P,1},\vv{\mbE,\mb{e}})\notag\\
&&                           & -Pr(D=1|P=0,\vv{\mbE,\mb{e}}) \label{eq:ce-e}
\end{align}

The discrimination score of $P\in \mb{P}$ in {\it data set $r$} is the E-group size weighted average:
\begin{align}
   \delta(P, r) & =  \sum_{e\in stra(\mb{E},r)} \delta(P,e)*|e|/|r| \label{eq:ce-r}
\end{align}
where $e$ is overloaded to also represent the signature $\mb{E}=\mb{e}$ of $e$ in Formula \bref{eq:ce-e}. Obviously $\delta(P, r)\le max_{{e\in stra(\mb{E},r)}} \{\delta(P, e)\}$ because of the average.

A data set has multiple protected variables. The discrimination score of a data set (without a specific $P$) is:
\begin{align}
   \delta(r) & =  \underset{P\in \mbP} {max} \ \delta(P, r) \label{eq:ce-PR}
\end{align}

\begin{definition}[Discrimination]
Given a data set $r$ and a user-defined discrimination score threshold $\alpha$, 
\begin{itemize}
 \item a protected group with a specific $P$ ($P\in \mb{P}$) is \bfit{group-discriminated} in the E-group $e$  if $|\delta(P, e)|> \alpha$;  

\item a protected group on $P$ is \bfit{globally discriminated} if $|\delta(P, r)|> \alpha$, and  

\item data set $r$ is \bfit{discriminatory} if $|\delta(r)|>\alpha$. $r$ is \bfit{discrimination-safe} if $|\delta(r)|\le\alpha$. $r$ is \bfit{discrimination-free} if $\delta(r)=0$. 
\end{itemize}
\end{definition}

\iffalse
\smallpara{The relationship between this definition with other definitions} The term discrimination score was used in \cite{zlio-review-2017,Calders2010} and is formulated as  $ds = Pr(D=1|P=1)-Pr(D=1|P=0)$ for a protected attribute $P$ over a data set. It is also called risk difference \cite{Ruggieri2014} and selection lift \cite{Pedreschi2009}. This metric has been widely used \cite{zlio-review-2017}. Our potential outcome based discrimination metric subsumes the risk difference measure in that if no explanatory attributes are specified, our metric produces the same score as that of risk difference. 

The potential outcome based score is semantic enabled and consequently more powerful. The explanatory variables in our metric ensure that the comparison of the protected group with the non-protected group is in the same application context. For example if profession is set as an explanatory attribute, our metric ensures that the income of female surgeons be compared with that of male surgeons. It guarantees no comparison will be done between the income of female surgeons with that of male kitchen hands. 
\fi

\smallsubsec{The problem}
Consider a data set $r$ on schema $R=\{D\}\cup\mb{P}\cup\mb{E}\cup\mb{S}$ and a classifier  ${\cal M}$. For each tuple $t\in r$, ${\cal M}(t)$ gives a prediction $\hat{D}_t$.  Let $\hat{r}$ be the data set from the predictions of ${\cal M}$: $\hat{r}=\{<t,\hat{D}_t>|t\in r \land \hat{D}_t={\cal M}(t)\}$.  As shown later on by the experiments, $\hat{r}$ is mostly discriminatory with regard to the predicted outcome $\hat{D}$ both at  the group level and globally.    

Our problem is whether we can design a method to build a model  $\Psi$ so that for each tuple $t\in r$ and its prediction $\hat{D}_t={\cal M}(t)$,  $\Psi(t, \hat{D}_t)$ also gives a prediction $D^\diamond_t$, and the data set $r^\diamond$ from the predictions of $\Psi$, $r^\diamond=\{<t,\hat{D}_t,D^\diamond_t>|t\in r \land \hat{D}_t={\cal M}(t) \land D^\diamond=\Psi(t,\hat{D}_t)\}$ is non-discriminatory with regard to the predictions $D^\diamond$ at both the group level and globally for all protected attributes. 

The challenge of the problem is that $\Psi$ must satisfy all discrimination constraints of multiple protected attributes in $\mb{P}$  so that every protected attribute is discrimination-safe after the adjustment. At the same time, the process should minimize the changes to the decision boundaries so that the predictions  $D^\diamond$  still have good accuracy. 

We note that we consider multiple individual protected attributes in $\Psi$, but the calculation of discrimination score $\delta(P,e)$ is still for each single protected attribute at a time. If the score needs to be computed for a combination of multiple protected attributes {\bf P'}, a new protected attribute $P^n$ should be created in data pre-processing to have the values of: $P^n=1$ if all the protected attributes in {\bf P'} have value 1 (e.g. black female) and $P^n=0$ otherwise  \cite{Friedler18-testbed}.

The related work to this problem is reviewed in the section of Related Work.    

\smallsubsec{Calculation of discrimination score}
The probabilities in Formula \bref{eq:ce-e} can be calculated in tuple counts of the DP-divisions in a data set as shown next. Given an E-group $e$ and a protected attribute $P$, \bfit{DP-divisions} are subsets of tuples with the same $D$ (outcome) value and the same $P$ (protected) value in $e$.  The concept of a division is a stratum of ${D}\cup{P}$, but it is used here to make the terminology distinct. $e$ thus has four DP-divisions because $D$ and $P$ are  binary. The tuple count of each division is denoted by $f_{ij}$ where the subscript $i$ means the $D$ value and the subscript $j$ is the $P$ value. For example, $f_{11}$ means the tuple count in the division $(D=1,P=1)$. The symbols denoting the counts are defined in Table \ref{tb-count-DP}, called the \bfit{counts table}. 

\begin{table}[h] \small\begin{center}
\caption{Tuple counts of DP-divisions		 \label{tb-count-DP}}
\begin{tabular}{|c |c c|c|} 
%  \multicolumn{4}{c}{(a) tuple counts  of training data} \\ 
  \hline
		   & $P=1$ & $P=0$ & sum \\ \hline
	$D=1$  &   $f_{11}$    &  $f_{10} $     &  $f_{1*} $   \\  
	$D=0$  &   $f_{01}$    &  $f_{00}$      &  $f_{0*} $  \\  \hline
	sum &        $f_{*1} $  &    $f_{*0}$  &  $f$ \\\hline
\end{tabular} 
\end{center}
\end{table}

With  DP-divisions, Formula \bref{eq:ce-e} can be calculated in Formula \bref{eq-count-dc}. Obviously each fraction in the formula is bounded by 1 and as a result, $\delta(P,e)$ is bounded to $[-1,1]$. 
\begin{align}
 & \delta(P,e) = \frac{f_{11}}{f_{11}+f_{01}}-\frac{f_{10}}{f_{10}+f_{00}}	\label{eq-count-dc} 
  \end{align}

In the special cases where there is no tuple in the contrast divisions for the protected attributes, i.e., $f_{11}=f_{01}=0$ or $f_{10}=f_{00}=0$, the discussion of discrimination in this case is not meaningful and no discrimination is possible. Then, $\delta(P,e)$ is defined to be 0.

\section{Removal of discrimination}

We describe our solution to removing discrimination from predictions in this section. We build a model $\Psi$ based on the training data $r$ and the predictions from a classifier ${\cal M}$. $\Psi$ then adjusts the prediction, i.e., flip the prediction, of a new instance to produce a new prediction $D^\diamond$ based on $(t,\calM(t))$.

\smallsubsec{Adjustment model $\Psi$}
Now we analyze the problem. Let $e$ be an E-group. For each $t\in e$, the model $\calM$ produces a prediction $\hat{D}_t$. Assume that the new data set for $e$ after the predictions is $\hat{e}=\{<\hat{D}_t,t>\}$ and the column name for $\hat{D}_t$ is $\hat{D}$. Now we sort the tuples in   $\hat{e}$ in descending order of values of the attributes $\hat{D},P_1,...,P_h,D$ where $P_1,...,P_h$ are all the protected attributes. From the sorted $\hat{e}$  we derive strata on $\hat{D},P_1,...,P_k,D$ and count the number of tuples (stratum size, denoted by $g$) in each stratum. In this way, we derive a table called a $\hat{D}\mbP D$-division table as shown in the left section of  Table \ref{tab:var-DPD}.  We note that the order among the $\hat{D}$, $\mbP$, and $D$  attributes matters. $\hat{D}$ must be to the left of all the $\mbP$ attributes, the target $D$ must be to the right of all the $\mbP$ attributes, and only $\mbP$ attributes can appear between $\hat{D}$ and $D$. The column order is important when the adjustments are formally modeled later on. The value in each row for $\hat{D}\mbP D$ is the signature of the $\hat{D}\mbP D$-division. 

We note that $\hat{D}\mbP D$-divisions are different from the previous presented DP-divisions. In DP-divisions, we consider only one target variable and one protected variable, while in $\hat{D}\mbP D$-divisions, we consider both the original target variable, the predicted target variable, and all protected variables.

\begin{table}[h] \begin{center}
\caption{$\hat{D}\mbP D$-divisions and variables for E-group $e$. $i$ is the division number. $\hat{D}={\cal M}(e)$ is the prediction. $P_a,P_b,P_c$ are protected variables. $D$ is outcome variable. $g$ is number of tuples in the $\hat{D}\mbP D$-div. $x$ is a variable indicating the number of tuples whose predicted value is to be adjusted in this div to make the whole set $e$ non-discriminatory. $lb$ and $ub$ are lower and the upper bounds of $x$. $u$ is a temporary variable indicating the sum of relevant x for each protected variable.
	  \label{tab:var-DPD}}
\includegraphics[scale=0.95]{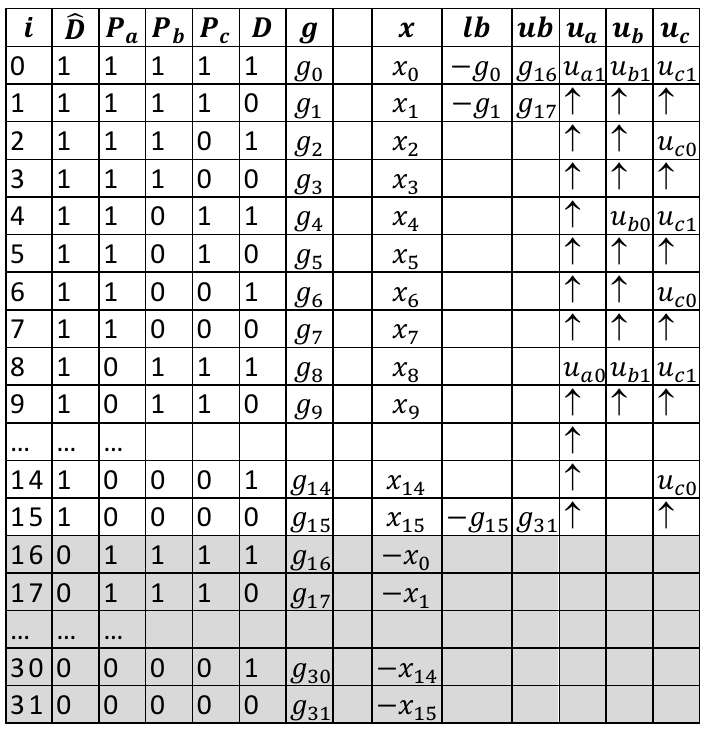}
\end{center}
\end{table}

Assume that after discrimination assessment, some predictions in $\hat{D}$ need to be adjusted to satisfy the discrimination requirements. There are a few points to be considered for the adjustment. The first one is the effect of adjusting a prediction. Consider the first row of Table \ref{tab:var-DPD} which is for the $\hat{D}\mbP D$-division of tuples with the signature $(\hat{D}=1,P_a=1,P_b=1,P_c=1,D=1)$. If we {\bf adjust (flip)} the prediction of a tuple in this division, $g_0$ should be reduced by 1 and $g_{16}$ should be increased by 1 because the tuple's prediction now is $\hat{D}=0$. $g_0$ and $g_{16}$ are {\bf counterparts} to each other because adjusting a tuple in division-16 causes the opposite change to the $g_0$ and $g_{16}$ values. Effectively, the adjustment of predictions means moving counts between the counterpart $\hat{D}\mbP D$-divisions. 

We examine the implication of prediction adjustments further. The discrimination score formula for $P_a$ is $\delta(P_a,e)=\frac{f^a_{11}}{ f^a_{11}+f^a_{01}}- \frac{f^a_{10}}{ f^a_{10}+f^a_{00}}$ where $f^a_{11}=\sum_{i=0..7} g_i$,  
$f^a_{01}=\sum_{i=16..23} g_i$, $f^a_{10}=\sum_{i=8..15} g_i$, and $f^a_{00}=\sum_{i=24..31} g_i$. adjusting the prediction of a tuple in div-2 means moving one(1) count from $g_2$ to $g_{18}$ and consequently moving one count from $f^a_{11}$ to $f^a_{01}$, making the first term of the discrimination score smaller. That is, the score adjusts to $\frac{f^a_{11}-1}{ f^a_{11}+f^a_{01}}- \frac{f^a_{10}}{ f^a_{10}+f^a_{00}}$. If we further adjust the prediction of a tuple in div-31, one(1) count is moved from $g_{31}$ to $g_{15}$ and the new score is $\frac{f^a_{11}-1}{ f^a_{11}+f^a_{01}}- \frac{f^a_{10}+1}{ f^a_{10}+f^a_{00}}$.

The motivation of adjusting predictions is to make final predictions non-discriminatory. Consequently the adjustments  should make all protected variables non-discriminatory at the same time. This is obviously a hard task because of multiple protected attributes. A adjustment may have impact on the discrimination score of one attribute, but not on another.  For example, consider $P_c$ in the table and the two adjustments in the previous paragraph, the new score is $\delta(P_c,e)_{new}=\frac{f^a_{11}}{ f^a_{11}+f^a_{01}}- \frac{f^a_{10}-1+1}{ f^a_{10}+f^a_{00}}$ which is the same as the score before the adjustment.
 
A straight forward adjustment is to make all predictions to one class. Then  discrimination vanishes. However, the utility of the model is completely lost. We see that in addition to the satisfaction of discrimination constraints, we also need to minimize the adjustment and make the predictions after adjustments as close as possible to the original class in $D$.

Next, we check two consecutive divisions having the same $\hat{D}\mbP$ value but different $D$ value such div-0 and div-1. If we adjust the prediction of a tuple in div-0, actually we are taking a wrong action as this would change an accurate prediction to a wrong prediction. However, if we adjust the prediction of a tuple in div-1, we are correcting a wrong prediction. So the adjustment should be discouraged if division has matching $\hat{D}$ and $D$ values, and encouraged if $\hat{D}$ and $D$ values do not match. So in adjusting predictions, in addition to satisfying discrimination constraints of all protected attributes, we also need to correct errors and to make as few new errors as possible.  

From all these discussion, we model the problem of adjusting predictions into an optimization problem as shown in Formulas \bref{eq-op-obj}-\bref{eq-op-u}.  We design variable $x_i$ to indicate the number of adjustments/flips to be made to the predictions of tuples in div-$i$. The variables $u$ are intermediate summary adjustments to the $f$ counts of each protected attribute.  A formal expression for each $u$ is complex and is described below. 

Let $m$ be the number of protected variables: $m=|\mbP|$, $p$ be the index of the protected variable $P_p$ in the $\hat{D}\mbP D$-division table, and $l_p=2^{m+1-p}$ be the number of consecutive divisions having the same $P_p$ value. In Table \ref{tab:var-DPD}, $P_a$ is the first protected attribute, so $p=1$ and $l_p=2^3=8$; $P_b$ is the second from the left, $p=2$ and $l_p=2^2=4$. In the $P_a$ column, we see eight consecutive 1's and eight consecutive 0's. In the same way, the $P_b$ column has four consecutive 1's and four consecutive 0's. Let $\iota_e(h)$ be a function returning 1 if $h$ is even and returning 0 if $h$ is odd, and let $\iota_o(h)$ return the  opposite to $\iota_e(h)$.  Thus, following Table \ref{tab:var-DPD}, the total number of adjustments made to $f^p_{11}$ is $u^p_1=\sum_{i=1..m'} x_i*\iota_e(i\tilde{\div} l_p)$, and the total number of adjustments made to $f^p_{10}$ is $u^p_0=\sum_{i=1..m'} x_i*\iota_o(i\tilde{\div} l_p)$ where $m'=2^{m+1}$ and $\tilde{\div}$ denotes the integer division.      
\begin{flalign}
      \begin{aligned}
	&&  min & \quad \sum_{i=1..m'} \frac{(x_{i}+g_{\iota_e(i)*m'+i})^2} {g_i+g_{m'+i}}  	\label{eq-op-obj} 
        \end{aligned}\ \\
    \begin{aligned}
	&& s.t. \ \  & for \ \ P_p \in \mbP 	: \\
	&&   &  |\frac{f^p_{11}+u^p_{1}}{f^p_{11}+f^p_{01}}   -
	           \frac{f^p_{10}+u^p_{0}}{f^p_{10}+f^p_{00}}|\le \alpha     \label{eq-op-alpha} 
        \end{aligned}\ \\
        \begin{aligned}
	 && & u^p_{1} = \sum_{i=1..m'} x_i*\iota_e(i\tilde{\div} l_p)  \\
	 && &  u^p_{0} = \sum_{i=1..m'} x_i*\iota_o(i\tilde{\div} l_p)  \label{eq-op-u} \\
	 && & bounds\ of\ x\ is\ in\ Table\ \ref{tab:var-DPD} \
        \end{aligned}
\end{flalign}

Formula \bref{eq-op-alpha} is the discrimination constraint for attribute $P_p$. In the formula, $u^p_1$ and $u^p_0$ are the summary adjustments to the $f$ counts to be decided via values of $x$. Normally, there are $m$ such constraints. If $f^p_{10}+f^p_{00}=0$ or $f^p_{11}+f^p_{01}=0$, the constraint is not necessary (see Formula \bref{eq-count-dc}'s comment). 

The objective function (Formula \bref{eq-op-obj}) takes some explanation. For div-0 in Table \ref{tab:var-DPD}, moving in a tuple means the overall error rate of prediction is reduced, but for div-1, moving a tuple out reduces the error rate.  Consequently for $\hat{D}=D$ divisions, a positive $x$ value increases correct predictions in their own division and reduces the errors in  its counterpart division $g_{\iota_e(i)*m'+i}$; for $\hat{D}!=D$ divisions, a negative $x$ value reduces errors in their own division $g_{i}$ and increases correct predictions in its counterpart. The square exponent ensures that no errors will be canceled and the divisor normalizes the terms.  

We allow $x$'s to take real values as the values are to be transferred to a probability and does not need to be integers. As a result of this, the optimization problem is a nonlinear problem with linear constraints and can be solved easily and efficiently using the $fmincon$ package in Matlab.  The following lemma guarantees that a solution exists for the problem.

\vspace{-.5ex}
\begin{lemma}	
This problem always has a solution. 
\end{lemma}
\vspace{-.5ex}
The lemma is true because if we let $u^p_1=-f^p_{11}$ and $u^p_0=-f^p_{10}$, the constraints \bref{eq-op-alpha} and \bref{eq-op-u} are satisfied although the objective function may be large.

Once the solution is obtained, a hash table
\begin{align} 
H = H(\hat{D}\mbP: <g,x>) \label{eq:hashmap}
\end{align} 
is built. The hash key $\hat{D}\mbP$ is comprised of the $\hat{D}$ column and all the $\mbP$ columns from the $\hat{D}\mbP D$-division table. For each key value, there are two rows with different $D$ values in the $\hat{D}\mbP D$-division table and $g$  is the sum of the $g_i$ values and $x$ is the sum of the $x$ values of the two rows. For example, the key $\hat{D}\mbP=<1,1,1,1>$ in Table \ref{tab:var-DPD} covers the first two rows. The value of the hash entry is derived from these two rows for the hash key, and the hash value is $g=g_0+g_1$ and $x=x_0+x_1$. The complete hash entry is $<1,1,1,1>:<g_0+g_1, x_0+x_1>$. 

This hash table $H$ forms the basis of our adjustment model. Then all hash tables of all E-groups is our adjustment model $\Psi$ which is also a hash table. The steps for constructing the model is summarized in Algorithm \ref{alg:fairmodel} called {\bf FairMod}.

\begin{algorithm}[th]
\caption{Fair Model $FairMod$}
\label{alg:fairmodel}
{\bf Input:}  data set $r$ over $\mb{R}=\mbO\mbE\mbP D$, trained prediction model $\calM$ from $r$, a discrimination threshold $\alpha$,\\
{\bf Output:} adjustment model $\Psi$  
\begin{algorithmic}[1]
\STATE $\Psi=\{\}$ // hash: key=E-group.sig, value=Hash.
\STATE $stra=stra(r, \mbE)$ // strata by $\mbE$
\FOR {each $e$ in $stra$}
  \STATE $H=\{\}$. // hash: key=$\hat{D}\mbP$, value=$<g,x>$
  \STATE $\hat{D}=\calM(r)$
  \STATE create $\hat{D}\mbP D$-division table. Sec 4.1
  \STATE construct optimization problem Eq. \bref{eq-op-obj}-\bref{eq-op-u}
  \STATE solve the problem to get solution $x$
  \STATE for each $\hat{D}\mbP$-division signature $sig$: add $(sig,<g,x>)$ to $H$ following Eq. \bref{eq:hashmap}   
  \STATE add $sig:H$ to $\Psi$ 
\ENDFOR 
\STATE return $\Psi$
\end{algorithmic}
\end{algorithm}

\vspace{.5ex}
  \smallpara{Prediction adjustment} \ \\
The hash table is used to decide prediction adjustments as the following. Given a new instance $t$, let the prediction for $t$ by model $\calM()$ be $\hat{D}_t = \calM(t)$. Then we retrieve the hash table as $<g_t,x_t>=H(\hat{D}_t t[\mbP])$ to get $g$ and $x$ values.
To decide the prediction, we generate a random number $rd$ in [0,1]. If $x_t<0  \land rd<x_t/g_t$, $D^\diamond_t = flip(\hat{D}_t)$; else  $D^\diamond_t = \hat{D}_t$ where $flip()$ is a function change a binary value to its opposite, and $D^\diamond_t$ is the adjusted prediction. Adjustments are only made to the divisions having a negative $x$ value which means moving out.   

\iffalse
The algorithm for our method is given in Algorithm \ref{alg:fairpred}. In the algorithm, the random number is moderated by a $(-log(x/g))$. This division is used to produce a smaller number so that movement can get a chance when the total size $g$ is small. Other ways of moderation may still be possible. 

\begin{algorithm}[th]
\caption{$FairMod$ in Prediction}
\label{alg:fairpred}
{\bf Input:}  prediction model $\calM$, adjustment model $\Psi$, new instance $t$ over $\mbO\mbE\mbP D$\\
{\bf Output:} prediction $D^\circ_t$  
\begin{algorithmic}[1]
  \STATE $\hat{D}_t=\calM(t)$;  \hspace{3em} $e.sig=t[\mbE]$;  
  \STATE $H=\Psi(e.sig)$; \hspace{2em} $<g,x,v>=H(<\hat{D}_t, t[\mbP]>)$
  \STATE $ rd= random(0,1)/(-log(x/g))$
  \STATE if $x<0 \land v<x \land rd < x/g$: \ \  $D^\circ_t=flip(\hat{D})$
  \STATE else: $D^\circ_t=\hat{D}$
\STATE return $D^\circ_t$
\end{algorithmic}
\end{algorithm}
\fi

\section {Experiments}
In this section, we present an evaluation of our method. The evaluation is to demonstrate the following points: (1) the effectiveness of our method for different prediction models and different data sets. (2) a comparison of our method with other methods.  (3)  the effect of different objective functions.

\smallpara{Data sets}
We use four real world data sets as shown in the following list. Name, size, source, and attributes of the data sets are shown. All data sets are processed to have binary (0,1) values. The values of ordinal attributes are binary-zed using median. Categorical attributes are binary-zed by taking majority and the rest.  The labels (P), (E), and (D) against some attributes indicate the types protected, explanatory, and outcome respectively. The attributes without a label are O-attributes.   

\begin{itemize}\small
             \topsep=1mm  \parsep=0mm \itemsep=0mm \parskip=0cm \itemindent=2em
\item[{\bf Adult}] US Census 1994. numb(rows)=48842; minority class=.25; 
     https://archive.ics.uci.edu/ml/data sets/adult \\
    Attributes: age45(P), natCountryUS(P), raceBlack(P), sexM(P), workPrivate(E), occuProf(E), workhour30(E), eduUni(E), relaNoFamily, married, income50K(D)

\item[{\bf Cana}] Canada Census 2011 \cite{canada-census} \footnote{The author wishes to acknowledge the statistical office that provided the underlying data making this research possible: Statistics Canada}. numb(rows)=691788; minority class=.34;  
     https://international.ipums.org \\
      Attributes: weight100(P), age50(P), sexM(P), edUni(E), occProf(E)\{\}, occSkilled(E), occOther(E), hoursfull(E), govJob, classSalary, income45K(D)
      
\item[{\bf Germ}] German Credit. numb(rows)=1000; minority class=.3;  
         https://archive.ics.uci.edu/ml/data sets/statlog+(german+credit+data) \\
         Attributes: age35(P), single(P), foreign(P), chkAccBal(E), duration20m(E), creditHistGood(E), purposeCar(E), credit2320(E), savings500(E), emp4y(E), installPct3(E), sexM(E), guarantor(E), resid3y(E), propertyYes(E), instPlanNon(E), houseOwn(E), creditAcc(E), jobSkilled(E), people2(E), hasTel(E), approved(D)

\item[{\bf Recid}]  Recidivate-violent \cite{larson2016-analyze-compas}; numb(rows)=4744; minority class=.14; 
	https://github.com/propublica/compas-analysis\\
   Attributes: sexM(P), age30(P), raceAfrica(P), raceWhite(P), raceOther(P), juvFelonyCnt1(E), juvMisdCnt1(E), juvOthcnt1(E), priorsCnt3(E), cjail1Month(E), cChargeisdemM(E), isRecid(E), score8(D) \\
\end{itemize}
The Recidivism data set is followed from \cite{larson2016-analyze-compas}. The score8 column stores predictions from a system called COMPAS. The isRecid column stores whether the person re-committed a crime. We want to see if score8 values can be accurately re-predicted. 
 
 \vspace*{0.5ex}
We now present the results of experimental evaluation of our method.

\begin{figure}[h]
\center      
\includegraphics[scale=0.9]{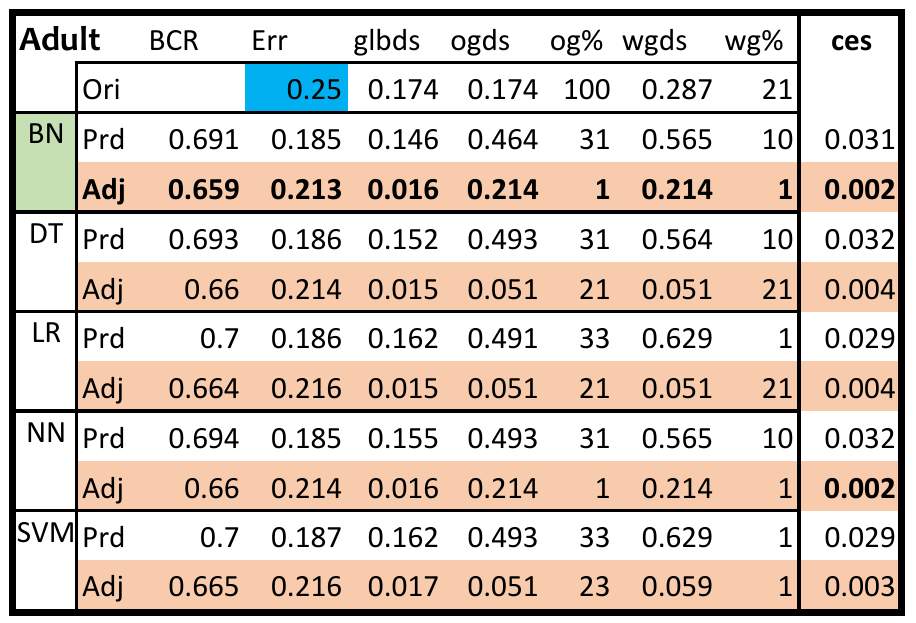}
\caption{Discrimination of Adult data set. $BCR$=balanced classification rate. $Err$=misclassification rate. $glbds$=global disc score. $ogds$=average score of over-limit E-groups. $og\%$=percentage of tuples in over-limit E-groups. $wgds$=max score of worst E-group. $wg\%$=percentage of tuples in worst E-group. $ces$=combined evaluation score. }
\label{plot:dsc-adult}
\end{figure}

\begin{figure}[h]
\center      
\includegraphics[scale=0.9]{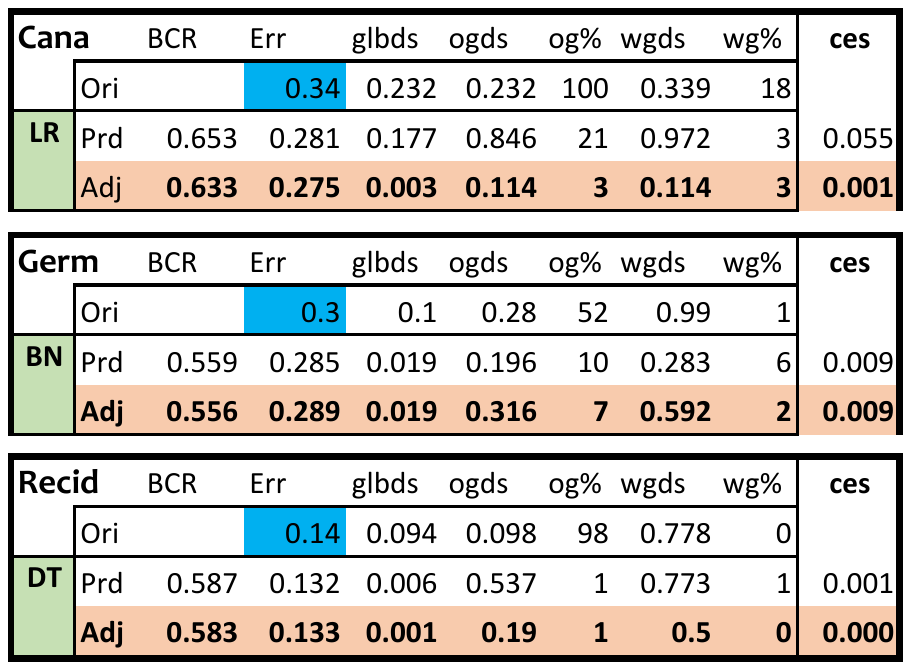}
\caption{Discrimination of Canada, German credit, Recidivate.  }
\label{plot:dsc-german}
\end{figure}

\smallpara{Effectiveness of the proposed method} To show the effectiveness, we choose 5 commonly used models, namely decision tree (DT), logistic regression (LR), Bayes network (BN), neural network (NN), and SVM, from SAS enterprise miner and run them on the four data sets described above. The data sets and their predictions are then input to our FairMod. The implementation outputs two rows of results: the first row contains discrimination scores for predictions before FairMod is applied and the second row contains the discrimination scores for the adjusted predictions after FairMod is applied.  Each row contains three discrimination scores:   the global discrimination score for all protected attributes ($glbds$), the average discrimination score for E-groups that are over the limit $\alpha$ ($ogds$), and the highest discrimination score of the worst E-group ($wgds$). The effectiveness results are shown in Fig.\ref{plot:dsc-adult}-\ref{plot:dsc-german}.

In the results, the discrimination scores for the original data set are also listed (row $Ori$ in the top). Row $Prd$ is for the scores before adjustment and row $Adj$ is for the scores after the adjustment. In the columns next to $ogds$ and $wgds$, we also list the percentage of tuples involved in the two cases ($og\%$ and $wg\%$). Accuracy of the predictions are measured by the balanced classification rate ($BCR$ - average of true positive and negative rates) which is an effective metric when class labels are not well balanced. $BCR$ and the misclassification rate ($Err$) are included for the inspection of adjustment cost. The combined evaluation score ($ces$) is defined as 
\[ces=(\frac{glbds+ogds*og\% + wgds*wg\%}{3}+Err)\div BCR\] 

We expect good effectiveness to have low discrimination scores, low misclassification rate, and high BCR. So, smaller evaluation score $ces$ is better.  

\begin{figure}[h]
\center      
\includegraphics[scale=0.4]{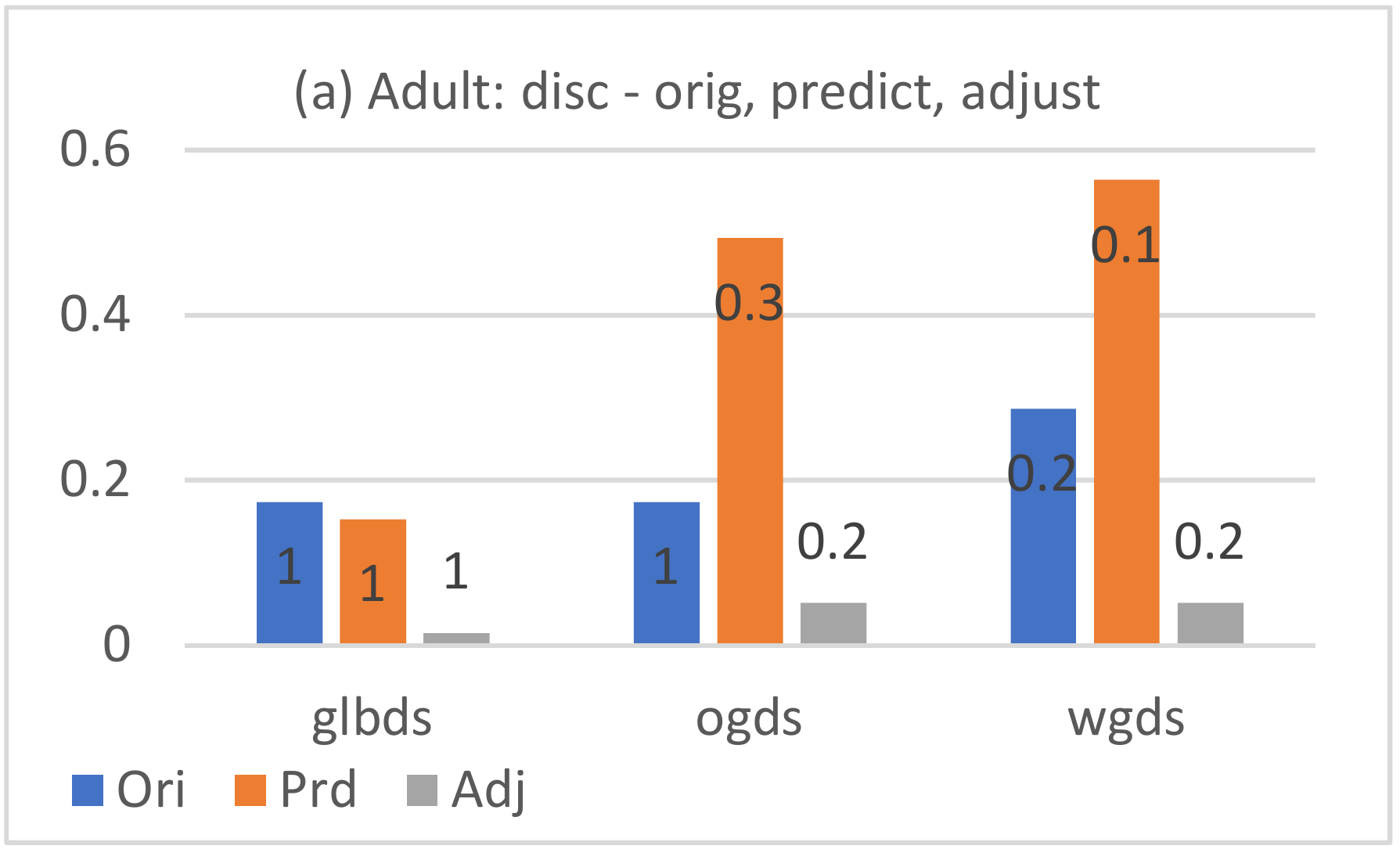}
\caption{Classification reduces global discrimination but increases E-group discrimination. The number on the bar is the percentage of tuples sharing the score. }
\label{plot:effect-classify}
\end{figure}

In the results for the Adult data in Fig.\ref{plot:dsc-adult}, classification itself reduces the global discrimination score from .174 to .146 but with the cost of increased score (.174 to .464) in over-limit E-groups (fortunately the percentage dropped from 100 to 31). This conclusion is also consistent in other data as shown in Fig.\ref{plot:effect-classify}.  

After the adjustment by $FairMod$, the final predictions are globally non-discriminatory (0.016$<\alpha$=0.05). The over-limit group score and the worst-group score both are higher than the threshold, but the respective percentages of tuples involved are very little. A close investigation found that when an E-group is small, the $g$ values of the $\hat{D}\mbP D$-divisions are small and the $x$ values are even smaller, leading to very few random numbers to be generated and to the fact that it is hard to say whether the generated numbers confirm to the even distribution.  

By comparing the two rows under `BN', the $ces$ value of $Adj$ is much smaller than that $Prd$, proving that the Adjustment is effective. Of course, the improvement in the discrimination score is with a cost (a drop of 0.032) of the balanced accuracy $BCR$ and an increased error rate (a rise of .028).  

For Adult and among all the models, the adjustment of `BN' is the most effective and highlighted in green in the front.  The cost of adjustment among all models is similar. 

Another observation is that a better classifier does not always leads to the best adjustment results. This is evidenced by comparing `LR' and `BN'. `LR' has the best accuracy, but did not win with regard to the $ces$ score. 

The same conclusions about adjustments on the Adult data set are also supported by the other three data sets (Fig.\ref{plot:dsc-german}) although the model that has the best adjustment results (marked green in front) is different in each data set.

\smallpara{Comparison of our method with existing methods}  We compare our method FairMod with two of the relevant methods CV \cite{Calders2010} and PR \cite{Kamishima2012} implemented in \cite{Friedler18-testbed}. Readers may reference \cite{Kamishima2018-actual-indep} for more comparisons. We use $Sex$ as the only protected variable in all three methods. The predictions from CV and PR are assessed using the metric in Eq.\bref{eq:ce-e} without an explanatory variable which meet the assumptions of CV and PR. 

With FairMod, we run the classifiers that best performed on each data set, set $Sex$ as the protected variable, and without setting any explanatory variable. The predictions from the classifiers and the training data then used in FairMod as input and adjusted predictions are output. The adjusted predictions are then assessed for discrimination as done for the CV an PR  methods.  

\begin{figure}[h]
\center      
\includegraphics[scale=0.9]{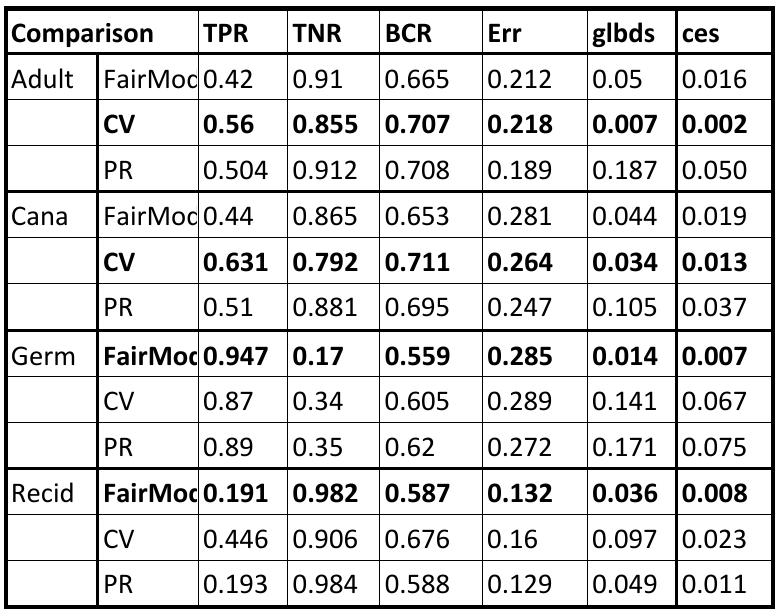}
\caption{FairMod is compared with CV and ROC }
\label{plot:comparison2cv}
\end{figure}

The results of comparison is shown in Fig.\ref{plot:comparison2cv}.  Here as CV and PR do not use explanatory variables, so only the global discrimination score ($glbds$) is valid to use. The combined evaluation score ($ces$) was adapted not to use $ogds$ and $wgds$. The results show that FairMod performed best in two out of the four data sets and CV best in the other two data sets.  When  Fig.\ref{plot:comparison2cv} is compared with Fig.\ref{plot:dsc-adult}-\ref{plot:dsc-german}, we found that both CV and PR have high BCR values, which means that on-the-shelf classifiers may need tuning to perform better.  The conclusion is that our method is better in the last two data sets and this is an addition to the extra power of handling multiple protected variables, explanatory variables and independence of classifier types.    

\smallpara{Effect of different objective functions} Equ.\bref{eq-op-obj} presented an objective function for the optimization problem and each term in the objective function is $\frac{(x_{i}+g_{\iota_e(i)*m'+i})^2} {g_i+g_{m'+i}}$. This function aims to minimize normalized misclassification count. We label the results of this objective function by $norm$. Actually there are two other possible objective functions. One is the function without the normalization factor. That is, we optimize the sum of misclassification count $(x_{i}+g_{\iota_e(i)*m'+i})^2$. This objective function is labeled with `Errc' for error count. The third function is the sum of  change $(x_{i})^2$ which aims to minimize the change to the original predictions. This function is labelled with `chg'.  We would like to know which function is the best to use. 

The results (details omitted) from the four data sets show that the three objective functions performed highly similar. 

\iffalse
\begin{figure}[h]
\center      
\includegraphics[scale=0.9]{dif-objs.pdf}
\caption{Effect of different objective functions}
\label{plot:dif-obj}
\end{figure}
\fi

 \section{Related Work}
Algorithmic discrimination has attracted a lot of research effort. The work focused in two areas: discrimination detection and discrimination removal from data and from models. The work on removal are in three categories: pre-processing, manipulation of model learning algorithms and post processing. We now review the work done in these directions.
  
\smallpara{Discrimination detection in data} The core problem of detection is to define and choose metrics to measure discrimination. \citet{zlio-review-2017} has a good summary of previous metrics \cite{Pedreschi2008,Pedreschi2009,Zliobaite2011,Fukuchi2013,Ristanoski2013-imbalancedData,Ruggieri2014,Feldman2015,Fish2016}.  Some recently proposed causality-based metrics are \cite{Fish2016,Kleinberg2016,wu2016c,Zhang2016,Li-bigdata2017,ZhangLu2016dr} and the metrics for individual discrimination are \cite{Luong2011,Dwork2012,Mancuhan2014,Zhang2016}.

\smallpara{Removal of discrimination from training data (pre-processing)}  \citet{Feldman2015} proposed to transform data in a data set so that the red-line attributes, those that correlated to the target, become independent to the target variable. \citet{Friedler18-testbed} summarized some of the work in this direction.

\smallpara{Modification of model training algorithms} \citet{Kamiran2010} proposed to combine information gain and discrimination gain in {\it decision tree} learning and to use a post-relabelling process to remove discrimination from predictions.  \citet{Calders2010} adjusts the probability in naive Bayes methods so that the predictions are discrimination free.   
\citet{Kamishima2018-actual-indep} proposed a regularization method for logistic regression method.  \citet{Zafar2015}  represented discrimination constraints via a convex relaxation and optimized the accuracy and discrimination in the SVM learning algorithm. 
\citet{Woodworth2017} proposes a two step method to build a non-discriminatory classifier. The data set is divided into two subsets S1 and S2. In the first step, a classifier is built to minimize error rate under the constraints of discrimination in data set S1. In the second step, a post-processing model is built on data set S2.          
 \citet{edward18-continuous-T} proposed a discrimination-aware measure for decision tree induction for continuous data. 
\citet{landerio16aaai-undertraining} proposes to use a weight under-training method by strengthening confounder features to build  a model. 
 \citet{Kearns2018-subgroupfairness} used an optimization method in model learning.
\citet{Kamishima2018-actual-indep} has done a deep analysis of CV2NB, ROC and proposed a method called universal ROC. 

\smallpara{Removal of discrimination from predictions of a model (post-processing)}
The work of \citet{Kamiran2010} relabels the predictions of leaf nodes of a decision tree to achieve discrimination goal. The post-processing model of Step 2 in \cite{Woodworth2017} minimizes the discrimination using the target, the predicted target and the protected variables on the second half of the training data.   
   \citet{Hardt2016} uses equalized odds to build a model for post-prediction manipulation. 
   \citet{Kamishima2018-actual-indep} proposed a method called universal ROC.

\section{Conclusion}
In this paper, we proposed a post-processing method to adjust predictions from a prediction model so that the predictions after the adjustment are non-discriminatory. The method allows the specification of explanatory attributes in discrimination assessment so that the context of discrimination can be specified and a CEO's income is not compared with that of an  employee. The method also allows the consideration of multiple protected attributes together so that the adjustment made to the predictions can satisfy discrimination requirements of all protected attributes. The method does not rely on a specific classifier, making it suitable in all applications. The comparison of the method with other existing methods shows that the proposed model is as effective as and better in some cases than existing methods.

\bibliographystyle{ACM-Reference-Format}
\bibliography{refs}

\end{document}